\documentclass{article}

\usepackage{arxiv}

\usepackage[utf8]{inputenc} 
\usepackage[T1]{fontenc}    
\usepackage{hyperref}       
\usepackage{url}            
\usepackage{booktabs}       
\usepackage{amsfonts}       
\usepackage{nicefrac}       
\usepackage{microtype}      
\usepackage{amssymb}
\usepackage{mathtools}
\usepackage{booktabs}
\usepackage{multirow,bigdelim}
\usepackage[square,numbers]{natbib}
\usepackage{pifont}
\usepackage{subfigure}

\title{Multi-Objective Hyperparameter Tuning and Feature Selection using Filter Ensembles}

\author{Martin Binder*\\
Ludwig-Maximilians-Universität München\\
\texttt{martin.binder@stat.uni-muenchen.de}
\And
Julia Moosbauer*
\\
Ludwig-Maximilians-Universität München\\
\texttt{julia.moosbauer@stat.uni-muenchen.de}
\And
Janek Thomas\\
Fraunhofer Institute for Integrated Circuits IIS\\
\texttt{janek.thomas@scs.fraunhofer.de}
\And
Bernd Bischl\\
Ludwig-Maximilians-Universität München\\
\texttt{bernd.bischl@stat.uni-muenchen.de}
}

\begin{document}
\maketitle
\newcommand\nnfootnote[1]{%
  \begin{NoHyper}
  \renewcommand\thefootnote{*}\footnote{#1}%
  \addtocounter{footnote}{-1}%
  \end{NoHyper}
}
\nnfootnote{These authors contributed equally to this work.}
\renewcommand*{\thefootnote}{\arabic{footnote}}

\begin{abstract}
Both feature selection and hyperparameter tuning are key tasks in machine learning. Hyperparameter tuning is often useful to increase model performance, while feature selection is undertaken to attain sparse models. Sparsity may yield better model interpretability and lower cost of data acquisition, data handling and model inference. While sparsity may have a beneficial or detrimental effect on predictive performance, a small drop in performance may be acceptable in return for a substantial gain in sparseness. We therefore treat feature selection as a multi-objective optimization task. We perform hyperparameter tuning and feature selection simultaneously because the choice of features of a model may influence what hyperparameters perform well.

We present, benchmark, and compare two different approaches for multi-objective joint hyperparameter optimization and feature selection: The first uses multi-objective model-based optimization. The second is an evolutionary NSGA-II-based wrapper approach to feature selection which incorporates specialized sampling, mutation and recombination operators. Both methods make use of parameterized filter ensembles.

While model-based optimization needs fewer objective evaluations to achieve good performance, it incurs computational overhead compared to the NSGA-II, so the preferred choice depends on the cost of evaluating a model on given data.
\end{abstract}


\maketitle

\section{Introduction}

Machine learning models often need to satisfy multiple objectives simultaneously to accomodate the nature of a practical setting.
Usually the main goal is predictive performance.
Especially on large and complex datasets this necessitates highly nonlinear algorithms, which have hyperparameters that need to be chosen carefully.
\emph{Hyperparameter optimization} poses a substantial challenge in machine learning: Besides few model-specific methods~\citep{franceschi2018bilevel,liu2018darts}, there are no general analytic representations of model performance w.r.t.\ hyperparameter settings. Performance therefore needs to be estimated using test-set evaluation or cross-validation.
Hyperparameter optimization is therefore an expensive black-box optimization problem.

Besides predictive performance, model sparsity is frequently another desirable objective.
According to \citet{Guyon2003}, sparser models help with interpretability, i.e.\ a better understanding of the underlying process that generated the data.
In addition to that, predictions can be made faster and more cost-effectively. Sparser models may even have better predictive performance, since they regularize against overfitting.

The process of \emph{feature selection} aims to select a small subset of relevant features while still constructing models with sufficient or even optimal predictive performance. 
There are two distinct model-agnostic approaches to feature selection~\citep{Guyon2003}: Filters and wrappers. 
\emph{Filters} use proxy measures to rank features by their estimated explanatory power, independently of the learning algorithm being employed. 
These include information theoretic measures, correlation measures, distance measures or consistency measures~\citep{Dash1997}.
In contrast, \emph{wrappers}~\citep{Kohavi1997} optimize model test-set performance directly over the space of feature subsets. Because every feature subset evaluation requires either one or multiple model fits, exhaustive search is usually infeasible, and a black-box discrete optimization search strategy is necessary. Commonly used are simple greedy methods like forward or backward search. More advanced methods like evolutionary algorithms can improve upon this~\citep{Xue2016}.
Because they directly optimize learner performance, wrappers often yield better results~\citep{Xue2016}. 

Feature selection is often considered as a single-objective task. Sometimes the feature selection step is only used to optimize performance~\citep{Kohavi1997}. However, in many applications it is desirable to forego a small drop in performance for a substantial gain in sparseness.
This leads to a natural treatment of the feature selection problem as a \textit{multi-objective optimization problem}: Maximize predictive performance while minimizing the number of features selected. 
Feature selection methods may aggregate model performance and number of features into a single objective function through a penalization term~\citep{Xue2016}. However, this implies a trade-off between performance and sparsity must be specified a-priori, which may be difficult.

Multi-objective optimization methods try to find a \emph{set} of solutions that represent different trade-offs between the different goals, enabling the user to consider the possible alternatives and to choose a fitting solution a-posteriori. This is beneficial for feature selection~\citep{XUE2013a}. 

Hyperparameter optimization and feature selection are often performed in separate steps.
We argue that jointly optimizing over the combined spaces of hyperparameters and feature subsets is beneficial and appropriate: The optimal choice of hyperparameter configuration is very likely to depend on the specific features that are included and vice-versa. Also, it is likely more computationally efficient to explore the joint spaces simultaneously.
When using the wrapper approach this combination is not trivial: The exponentially large binary search space of selected features now has to be fused with the mixed numeric-categorical space of hyperparameters. 

We present and adapt model-agnostic holistic approaches for both aspects discussed above: multi-objective and joint optimization of hyperparameters and feature sets. To guide the search, our methods make use of combinations of feature filter scores which are combined in a \emph{filter ensemble}.

Our approaches can be considered \emph{hybrid} filter-wrappers: they are fundamentally ``wrapper''-based, because they optimize model-performance, but also make use of filters.



\section{Related Work and Contributions}\label{sec:related}

In recent literature, \emph{Bayesian optimization} (``BO''), also referred to as \emph{model-based optimization}, has become a popular method for hyperparameter tuning of learning algorithms~\citep{Snoek2012}, often outperforming simple grid search or random search. Originally, BO generally relied on a Gaussian process (GP) model, such as in the Efficient Global Optimization (EGO) algorithm~\citep{Jones1998}. However, the GP does not typically scale well to high dimensions and large numbers of data points. Random forests have been proposed as an alternative surrogate model, as used in the popular \emph{SMAC} hyperparameter optimization method~\citep{Huttera}.

Automated machine learning (AutoML) deals with the configuration and optimization of complete machine learning pipelines, often encompassing data pre-processing, ML models, ensembling, and possibly post-processing steps. 
AutoML may encompass choosing among the many different possibilities of what method to use at each stage of the pipeline, optimizing the hyperparameters for these methods, and combining them in ways that yield well-performing ensembles.
Feature selection by filter or wrapper methods is an important pre-processing step already part of some AutoML frameworks.
Consequently, AutoML has the potential to jointly optimize hyperparameters and included features. 
\emph{autosklearn}~\citep{Feurer2015} for example integrates a filter-based feature selector, parameterized by a filter measure and a percentage indicating the fraction of highest-ranked features to be included.
Bayesian optimization is used to find the optimal filter (among a set of possible filters) and the best feature selection rate.

However, even though hyperparameter optimization and feature selection are both present in some AutoML frameworks, it is not their goal to find a good trade-off between predictive performance and sparseness.
Feature selection is merely used to improve predictive performance, without considering the preference for sparse models in light of better interpretability or other benefits.  
In fact, these frameworks often tend to produce considerably complex models. They may even introduce additional features through feature engineering in pursuit of increasing predictive performance as the only goal.
The results are often large and heterogeneous ensembles that are hard to interpret and deploy~\citep{pfisterer2019multi}.






Evolutionary algorithms are especially well suited for multi-objective optimization. 
According to the survey by \citet{Xue2016}, \emph{genetic algorithms} (GAs) are among the most commonly applied techniques for multi-objective feature selection.
Inspired by natural evolution, GAs apply recombination and mutation operators to iteratively improve the population of solution candidates. 
Through techniques like non-dominated sorting~\citep{Deb1995,Deb2002}, genetic algorithms have become a powerful tool for multi-objective optimization. 
Several GA-based methods have been proposed for the task of wrapper-based feature selection~\citep{Emmanouilidis,Hamdani2007,Waqas2009}.

There have been first investigations on simultaneous multi-objective hyperparameter optimization and feature selection using GAs: \citet{Bouraoui2018} proposed an SVM-wrapper approach based on the NSGA-II, using a shared representation of the feature configuration and algorithm hyperparameters. While their method of combining the search spaces is similar to our GA-based approach, \citet{Bouraoui2018} do not use specialized initialization and mutation operators that we have found to be necessary for good performance\footnote{See the Ablation Study in the Supplement, where their approach, ``Variant (1)'', is outperformed by all our methods on every dataset and learning algorithm with no exception.}. Another limitation is that their approach is not model agnostic but limited to SVMs only.

Numerous methods have been proposed for \emph{feature filtering}, and there are known ways of combining filters into \emph{filter ensembles}. These may rely on applying the same filter on diverse datasets (``homogeneous approach''), or different filters on the same dataset (``heterogeneous approach''), and there are different ways of aggregating the filter rankings~\citep{Bolon2018}. We extend the heterogeneous approach into a hybrid filter-wrapper selection method by optimizing a parameterized ranking aggregator.

There has been a lot of research on feature selection and hyperparameter optimization. However, we found that there is no \emph{general} algorithm or framework to perform model-agnostic simultaneous hyperparameter optimization and feature selection for predictive performance and model sparsity in a multi-objective fashion.
We therefore choose to tackle this problem from two directions. First we adapt a standard evolutionary multi-objective optimization method, the NSGA-II, for the particular problem of feature selection and hyperparameter tuning. We contrast this with a classical hyperparameter optimization algorithm, based on BO, extended to perform feature selection.

Our main contributions are: 
\begin{enumerate}
    \item We adapt the NSGA-II to the problem at hand by introducing specialized sampling and mutation operators that make use of filter ensembles to enhance optimization performance.
    \item We investigate how multi-objective Bayesian optimization (MOBO) methods can be used for this problem and propose an effective method that uses a filter ensemble for feature selection. 
    \item By conducting a benchmark of these approaches on a variety of tasks for different machine learning algorithms, we provide a comparison between our approaches and show how these algorithms perform compared to suitable baselines.
\end{enumerate}

\section{Problem Statement}\label{sec:problem}

We consider the machine learning problem of a given feature space $\mathcal{X}$ of vectors with $p$ components (``features''), an arbitrary outcome space $\mathcal{Y}$ (for example $\{-1, 1\}$ for a binary classification task), and a performance measure $L: \mathbb{R}^g \times \mathcal{Y} \rightarrow \mathbb{R}$ measuring the quality of predictions (in $\mathbb{R}^g$) given ground truth values. $g$ is 1 for regression tasks, and equal to the number possible outcome classes for classification. Data samples $\mathcal{D} = \left\{ \left(\mathbf{x}^{(1)}, y^{(1)}\right), \ldots, \left(\mathbf{x}^{(n)},  y^{(n)}\right)\right\} \in {\left(\mathcal{X} \times \mathcal{Y}\right)}^n$ are assumed to be $n$ i.i.d. realizations of random variables $(\boldsymbol{X}, \boldsymbol{Y})$ which follow a joint distribution $\mathbb{P}_{XY}$. A dataset is thus notably characterized by $n$, the number of samples available, and $p$, the number of \emph{features}.

Let $\mathcal{A}_{\boldsymbol{\lambda}, \boldsymbol{s}}$ be a learning algorithm that takes the given dataset $\mathcal{D}$ and constructs a model $f: \mathcal{X} \rightarrow \mathbb{R}^g$. The learning algorithm is parameterized by the \emph{feature configuration vector} $\boldsymbol{s} \in {\{0, 1\}}^p$, where $s_j = 1$ denotes that feature $j$ is included in the model. The sparseness of the resulting model is thus determined by the \emph{Hamming weight} of $\boldsymbol{s}$, i.e. the number of components of $\boldsymbol{s}$ that are 1. \emph{Hyperparameters} of the learning algorithm are summarized in a vector $\boldsymbol{\lambda} \in \Lambda$.  $\Lambda$ is a (possibly mixed) bounded space and may contain numeric, integer, and categorical values\footnote{Hierarchical dependencies of hyperparameters, e.g.\ kernel hyperparameters that appear only if a specific kernel is chosen, are a common extension but not considered in this work.}.



In general we are trying to construct a model $f=\mathcal{A}_{\boldsymbol{\lambda}, \boldsymbol{s}}(\mathcal{D})$ which minimizes the \emph{generalization error} $\textrm{GE}[f]=\mathbb{E}_{\mathbb{P}_{XY}}\left[L(f(\mathbf{x}), y)\right]$.  However, the generalization error can only be estimated using in-sample data $\widehat{\textrm{GE}}[\mathcal{A}_{\boldsymbol{\lambda}, \boldsymbol{s}}, \mathcal{D}]$ through a resampling technique such as cross-validation. 

This setup gives rise to the bi-objective hyperparameter optimization and feature selection problem:
\begin{equation*}
\min_{\boldsymbol{\lambda} \in \Lambda, \boldsymbol{s} \in {\{0, 1\}}^p} \left( \widehat{\textrm{GE}}\left[\mathcal{A}_{\boldsymbol{\lambda}, \boldsymbol{s}},\mathcal{D}\right], \sum_{i = 1}^p c_i s_i, \right)\textrm{.}
\end{equation*}
The setup regards estimated generalization error as one objective and the cost of features considered as another.

It is possible, and a trivial extension of our method, to consider arbitrary costs $c_i$ for each feature $i$. However, in our benchmarks we limit ourselves to equal costs $c_i=1/p$. The resulting measure corresponds to the \emph{fraction of selected features}, $\mathrm{ffrac}=\frac{1}{p}\sum_i s_i$, ranging from 0 to 1.


\section{Multi-Objective Hyperparameter Tuning and Feature Selection: Two Approaches}
\label{sec:methods}



Our two methods are based on two popular multi-objective optimization methods, adapted for the particular task of hyperparameter tuning and feature selection: A model-based approach, and an approach based on an evolutionary algorithm. Both approaches make use of feature filters to accelerate the search.

\subsection{Feature Filters and Filter Ensembles}\label{sec:filterens}

The number of possible feature configurations $\boldsymbol{s}$ is exponential in $p$, so a large number of performance evaluations would be necessary to explore the performance space. Therefore, both our optimization approaches make use of \emph{feature filters}. These are methods that heuristically score individual features according to their apparent relevance for the outcome variable. Because there are various methods to estimate this relevance, we consider a collection of $M$ feature filters, which generate scores ${F^m(\mathcal{D})}_j$, $m = 1, \ldots, M$, $j = 1, \ldots, p$ for each feature $j$ of a training dataset $\mathcal{D}$. Different methods may generate scores on different scales, but we rank-transform and scale these scores to values ranging in equi-distant steps from least (value of 0) to most (value of 1) relevant for the outcome variable.

We propose combining multiple filter methods into \emph{filter ensembles} similarly to \citet{Dittman2012}, but extending their method by using \emph{weighted} average rank aggregation. The ensemble filter score $\mathrm{EF}_j$ for feature $j$ is calculated as the weighted average according to weight vector $\boldsymbol{w} \in [0, 1]^M, \sum_{i} w_i = 1$:
\begin{equation}
\label{eqn:fw}
    \mathrm{EF}_j(\boldsymbol{w})=\sum_{m=1}^M w_m {F^m(\mathcal{D})}_j\textrm{.}
\end{equation}

The weighting parameter $\boldsymbol{w}$ can be optimized, extending the filter into a hybrid filter-wrapper approach.


\subsection{Bayesian Optimization Approach}

\emph{Bayesian optimization} (BO), also often referred to as \emph{sequential model-based optimization}, has been successfully used for machine learning hyperparameter optimization in many applications~\citep{Snoek2012}. The principle of BO is based on two steps, which are performed in turn. First, a so-called \emph{surrogate model} is fitted to model the relationship between decision variables (e.g. hyperparameter values) and the objective value (e.g. estimated generalization performance). The surrogate model generates cheap approximations of the (generally expensive to evaluate) objective function values. In a second step, an \emph{infill criterion} is used to find promising decision values to be evaluated on this expensive function. The infill criterion has to face a trade-off between ``exploitation''---evaluating points for which the surrogate predicts good performance---and ``exploration''---evaluating points where predictive uncertainty is high.

There are different ways of adapting BO to perform multi-objective Bayesian optimization (``MOBO'')~\citep{Horn2015}, which has been applied successfully to hyperparameter tuning~\citep{horn2016multi}. We chose the \emph{Parego} method~\citep{Knowles2006}. Parego scalarizes the different objectives through the Chebyshev norm using a random weight vector, which is sampled again for every point being proposed. Parego has the advantage over many other MOBO methods of only requiring a single surrogate model fit for a proposed point~\citep{horn2016multi}, but other MOBO approaches can trivially be used as a substitute within our method.

The number of possible feature configurations is exponential in $p$, so a large number of evaluations would be necessary to fit an accurate surrogate model on performance values. We therefore investigate the use of prior knowledge from \emph{feature filter} scores to reduce the dimensionality of the search space. Two possible methods for simultaneous use of multiple feature filter methods are considered (see also Figure~\ref{fig:indivexplanation}):

\begin{figure}
\raggedright
   \quad GA-MO-FE: NSGA-II with feature ensemble mutation \\
   \vspace{0.15em}
    \centering
    \includegraphics[width=0.4\textwidth,page=2]{figures/mosmafspaper_individuals_cropped} \\
    \vspace{0.1em}
\raggedright
\quad GA-MO: NSGA-II with Hamming-weight preserving mutation  \\
\vspace{0.15em}
    \centering
    \includegraphics[width=0.4\textwidth,page=1]{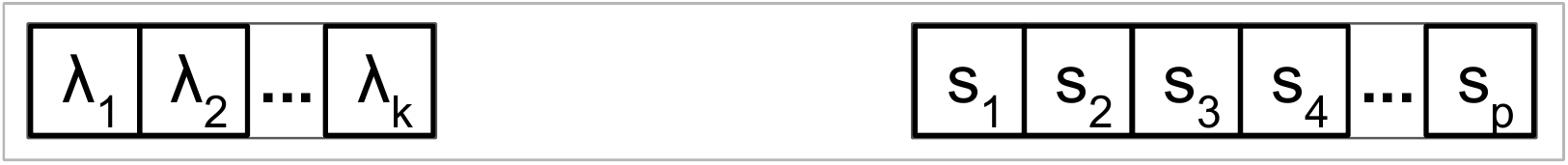} \\
    \vspace{0.1em}
\raggedright
\quad GA-MO-FE-NJ: NSGA-II for features; hyperparameters fixed\\
\vspace{0.15em}
    \centering
    \includegraphics[width=0.4\textwidth,page=3]{figures/mosmafspaper_individuals_cropped} \\
    \vspace{0.1em}
\raggedright
\quad BO-MO-FE, BO-SO-FE, BO-MO-FE-NJ: BO with filter ensemble \\
\vspace{0.15em}
    \centering
    \includegraphics[width=0.4\textwidth,page=5]{figures/mosmafspaper_individuals_cropped} \\
    \vspace{0.1em}
\raggedright
\quad BO-MO, BO-SO: BO with individual filter selection \\
\vspace{0.15em}
    \centering
    \includegraphics[width=0.4\textwidth,page=4]{figures/mosmafspaper_individuals_cropped} \\  \caption{Representation of individuals in different variants of our proposed optimization methods as described in Section~\ref{sec:algorithms}; $\mathbf{\lambda}$, $\mathbf{w}$, and $\mathbf{s}$ are vectors as described in Section~\ref{sec:methods}. 
    }
    \label{fig:indivexplanation}
\end{figure}

\begin{figure}
\centering
\includegraphics[width=0.6\textwidth]{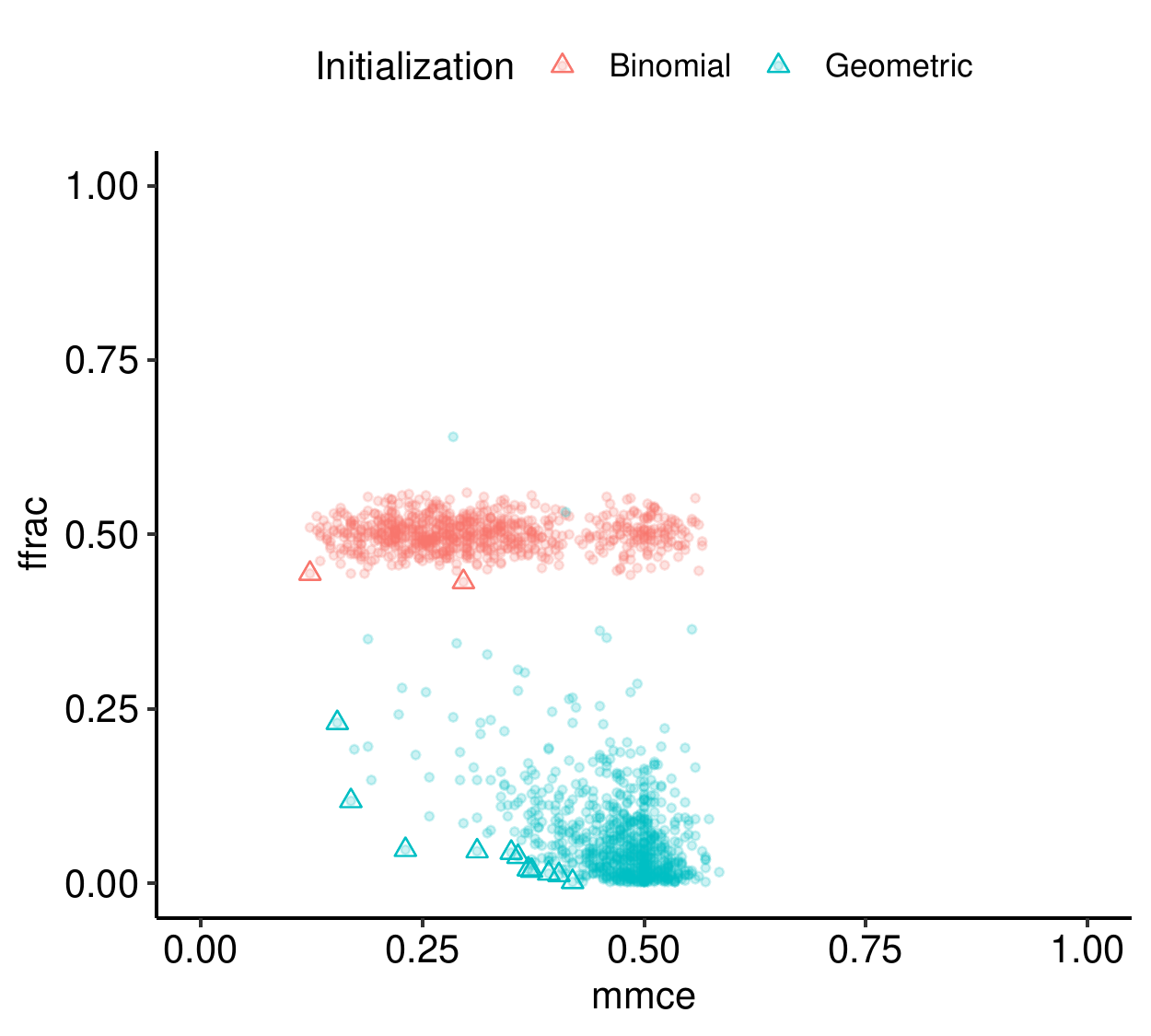} \\
\caption{Example population (800 samples) of \emph{xgboost} configurations randomly sampled and evaluated on the \emph{madelon} task. Red: $\frac{1}{2}$-Bernoulli sampling. Blue: Geometric initialization (Section~\ref{sec:geominit}). Triangles show non-dominated individuals. Although randomly initialized and not yet optimized, the blue individuals dominate a larger area. 
}
\label{fig:geominit}
\end{figure}

\subsubsection{Individual filter selection}\label{sec:bonofe} In this method, we introduce a discrete \emph{filter selection hyperparameter}~$m$, as well as a \emph{feature fraction hyperparameter}~$\mathrm{ffrac} \in [0, 1]$. For each model evaluation, only the most relevant $\lceil p\cdot{}\mathrm{ffrac} \rceil$ features, according to filter with index $m$, are included in the model. This approach is similar to the one taken in auto-sklearn~\citep{Feurer2015}, although there the feature selection problem was not considered as a multi-objective problem.

\subsubsection{Filter ensemble selection}\label{sec:bofe}
This method uses the filter ensemble as shown in Equation~\ref{eqn:fw} and introduces the vector $\boldsymbol{w}$, as well as the aforementioned feature fraction~$\mathrm{ffrac}$, as hyperparameters. The most relevant $\lceil p\cdot{}\mathrm{ffrac}  \rceil$ features, according to $\mathrm{EF}_j(\boldsymbol{w})$, are included in the model.

\subsection{Evolutionary Approach}
The \emph{Nondominated Sorting Genetic Algorithm II} (NSGA-II)~\citep{Deb2002a} is an evolutionary multi-objective algorithm that uses nondominated-sorting to preferably select individuals close to the Pareto-front of the problem. It iterates through generations of each a \emph{reproduction}, a \emph{crossover}, a \emph{mutation}, and a \emph{survival} step that generate the population of the next generation. 

GAs often represent individuals as vectors of binary, discrete, or continuous values, depending on the optimization problem. Because hyperparameters generally have various types, we use the Cartesian products of operators that operate in different ways on the various types, following \citet{Li2013}. This means e.g. that numeric hyperparameters undergo Gaussian mutation, while categorical hyperparameters undergo uniform mutation etc. Table~\ref{tab:recombmutation} summarizes the chosen recombination and mutation operators for the respective hyperparameter types. For the hyperparameter mutations, we use self-adapting step sizes and mutation probabilities as suggested by \citet{Li2013}.

The feature configuration vector parameter $\mathbf{s}$ plays a special role, because $\mathbf{s}$ maps to the objective of the fraction of selected features, which is being optimized, in a straightforward manner. For datasets with many features it makes up a large part of the search space under consideration compared to the other hyperparameters. The initialization and mutation performed on this parameter should therefore be considered in detail. 

\begin{table}
\centering
\caption{Summary of mutation and recombination parameters used in our GA-based algorithms. \emph{s.a.}: the parameter is controlled by self-adaption~\citep{Li2013}.
Filter-ensemble mutation of feature configurations is used in GA-MO-FE(-NJ), Hamming-weight preserving mutation in GA-MO.}
\begin{tabular}{l l p{3.2cm}}
\toprule
parameter & recombination                           &  mutation \\
\hline 
numeric        & SBX $(\eta = 5)$                        & Gaussian \newline ($p=0.1$, $\sigma^2$ = s.a.) \\
integer        & rounded SBX $(\eta = 5)$                & rounded Gaussian\newline  ($p=0.1$, $\sigma^2$ = s.a.) \\
categorical    & uniform ($p = 0.5$)    & uniform mut. ($p$ = s.a.) \\
binary         & uniform ($p = 0.5$)    & uniform mut. ($p$ = s.a.) \\
features       & uniform ($p = 0.5$)    & Filter-ensemble or\newline Hamming-weight pres. \\
\bottomrule
\end{tabular}
\label{tab:recombmutation}
\end{table}

\subsubsection{Geometric initialization}\label{sec:geominit} A naive approach for feature configuration initialization would be Bernoulli-sampling of each feature selection bit $s_i$ individually, possibly biased towards a low expected number of selected features to favor relatively sparse solutions~\citep{Bischla}. However, this gives rise to a binomial distribution of the number of selected features $\sum_i s_i$ with standard deviation $\sim O\left(\sqrt{p}\right)$. For even moderately large values of $p$, this fails to cover the objective space evenly along the dimension of the selected feature fraction, see Figure~\ref{fig:geominit}. We propose to sample values of $\boldsymbol{s}$ such that the sum of selected features covers the whole feature fraction objective. Therefore, we elect to use a truncated geometric distribution of number of included features to encode our preference for sparse models. This is achieved by sampling the desired number of included features $S$ as a truncated geometrically distributed random integer between $0$ and $p$, and then uniformly sampling from all vectors $\mathbf{s}$ that satisfy $\sum_i s_i=S$.

The method introduces the success probability of the geometric distribution as a configuration parameter. It can be set by the user to encode a relative preference for sparsity. We chose to use an empirically determined value by fitting decision trees on 100 random subsets of 90\% of the data set and determining the average number of distinct split variables.

\subsubsection{Filter-ensemble based initialization}\label{sec:gafi} It is possible to enhance the geometric initialization by including prior knowledge gained from feature filter methods. The goal is to select the features that have high filter scores with larger probability than the ones with lower ranking, while still having an approximately geometric distribution over the number of total features selected. We make the initial distribution of bit $j$ dependent on the filter ensemble value as described in equation~\ref{eqn:fw}, with $\boldsymbol{w}$ uniformly randomly sampled from the simplex $\boldsymbol{w} \in {\left[0, 1\right]}^M$, $\sum_{m=1}^M w_m=1$. For each individual to be initialized, we first sample $S$ as in \ref{sec:geominit}. Each bit $j$ is then sampled from a Bernoulli-distribution with parameter:
\begin{equation}
\label{eqn:filtersample}
\pi_{\mathrm{B}}(\boldsymbol{w}, S) = \frac{\mathrm{EF}_j(\boldsymbol{w})\left(S+1\right)}{\mathrm{EF}_j(\boldsymbol{w}) S + \left(1 - \mathrm{EF}_j(\boldsymbol{w})\right)\left(p - S\right) + 1}\textrm{.}
\end{equation}

\subsubsection{Hamming-weight preserving mutation}\label{sec:ganofe} Performing random bit-flip mutation on the feature selection vector entails similar problems to Bernoulli-initialization: a bit-flip with probability $\pi_{\mathrm{Mut}_\mathrm{bin}}$ is equivalent to erasing a bit with probability $2\pi_{\mathrm{Mut}_\mathrm{bin}}$ and sampling it anew from a $\frac{1}{2}$-Bernoulli distribution. This biases the mutation result towards the $\frac{1}{2}$ point of the feature fraction objective. Instead, we choose to approximately preserve the Hamming weight. This is done by sampling erased bits from a Bernoulli-distribution with parameter $\pi_{\mathrm{B}}=(S+1)/(p+2)$, where $S = \sum_i s_i$ is the Hamming weight of the original vector.

\subsubsection{Filter-ensemble based mutation}\label{sec:gafe}
Just as for initialization, mutation can be made dependent on feature filter ensemble ranks to preferentially include more relevant features. As in the Hamming-weight preserving mutation, each bit is erased with a probability of $2\pi_{\mathrm{Mut}_\mathrm{bin}}$ and then drawn from a Bernoulli-distribution. The parameter is as in equation~\ref{eqn:filtersample} for filter-based initialization, with $S$ the Hamming weight of the original vector, and $\boldsymbol{w}$ a weighting vector. This weighting vector itself becomes part of the search space to achieve self-adaption, as described in \citet{Li2013}, which results in $\mathbf{w}$ being optimized during the NSGA-II run. This feature configuration mutation method is (approximately) Hamming-weight preserving.

\subsection{Implementation and Reproducibility}

All proposed methods are implemented and publicly accessible through the \texttt{mosmafs} \texttt{R} package published on CRAN%
\footnote{\url{https://CRAN.R-project.org/package=mosmafs}}.
For full reproducibility of the results we publish the code used to perform the benchmark experiments on Github%
\footnote{\url{https://github.com/compstat-lmu/mosmafs/tree/master}}. 


\section{Experiments}\label{sec:experiments}

We conduct experiments to answer the following research questions empirically:

\begin{enumerate}
    \item \label{enum:gavsbo} \emph{Evolutionary vs. Bayesian Optimization}: How do the proposed methods---the evolutionary and the Bayesian optimization approach---perform relative to each other? 
    \item \label{enum:fevsno}\emph{Effect of Filter Ensembles}: Do the methods benefit from using \emph{filter ensembles}?
    \item \label{enum:movsso}\emph{Multi-Objective vs. Single-Objective}: Does multi-objective optimization find much sparser solutions without a major loss in predictive performance compared to single-objective optimization? 
    \item \label{enum:jvsnj}\emph{Simultaneous Hyperparameter Tuning and Feature Selection}: Is it beneficial to perform hyperparameter optimization and feature selection \emph{simultaneously} compared to performing the tasks sequentially? 
\end{enumerate}


\subsection{Benchmark Datasets}
\label{sec:tasks}

We consider real-world binary classification tasks that are publicly accessible through the OpenML platform~\citep{Vanschoren2014} (see Table~\ref{tab:datasets}). To eliminate algorithmic factors that might influence the result (like class imbalance correction, feature encoding, or handling of missing values), we included datasets fulfilling the following criteria: (roughly) balanced outcome classes, a purely numeric feature space, and no missing values. Further, datasets have been chosen to represent a large variety in terms of dimensionality ($30 \le p \le 10937$)\footnote{To limit the computational resources necessary for our experiments we did not investigate even larger datasets.} and in terms of instances per dimension ($0.02 \le n / p \le 18.97$).

Although our method can be used for regression as well as classification tasks, we did not perform experiments on the former. 

\begin{table}
\caption{Description of the datasets being used. $n$ denotes the number of observations and $p$ the total number of features. The class ratio gives the proportional size of the smaller outcome class. The dataset id (did) is the unique identifier for the dataset on the OpenML platform~\citep{Vanschoren2014}.
}
\label{tab:datasets}
\centering
\begin{tabular}{lrrrrr}
\hline
\toprule
name & $n$ & $p$ & class ratio & $n / p$ & did \\ \hline
wdbc            &  $569$ &    $30$ & $0.37$ & $18.97$ & $1510$ \\ 
ionosphere      &  $351$ &    $33$ & $0.36$ & $10.64$ & $59$ \\ 
sonar           &  $208$ &    $60$ & $0.47$ &  $3.47$ & $40$  \\ 
hill-valley     & $1212$ &   $100$ & $0.50$ & $12.12$ & $1479$ \\ 
tecator         &  $240$ &   $124$ & $0.43$ &  $1.94$ & $851$ \\ 
semeion         &  $319$ &   $256$ & $0.50$ &  $1.25$ & $41973$ \\ 
madeline        & $3140$ &   $259$ & $0.50$ & $12.12$ & $41144$\\ 
lsvt            &  $126$ &   $307$ & $0.33$ &  $0.41$  & $1484$ \\ 
madelon         & $2600$ &   $500$ & $0.50$ &  $5.20$ & $1485$\\ 
isolet          &  $600$ &   $617$ & $0.50$ &  $0.97$ & $41966$  \\ 
cnae-9          &  $240$ &   $282$ & $0.50$ &  $0.85$  & $41967$ \\ 
arcene          &  $200$ &  $9961$ & $0.44$ &  $0.02$ & $1458$\\ 
AP\_Breast\_Colon &  $630$ & $10935$ & $0.45$ &  $0.06$ & $1145$ \\ 
AP\_Colon\_Kidney &  $546$ & $10935$ & $0.48$ &  $0.05$ & $1137$ \\ 
\bottomrule
\end{tabular}  

\end{table}

\subsection{Learning Algorithms and their Hyperparameters}
\label{sec:algos}

We consider three different classifiers that are tuned by the optimization algorithms proposed in Section \ref{sec:methods}: The support vector machine classifier (SVM)~\citep{Cortes1995} with Gaussian kernel, extreme gradient boosting (xgboost)~\citep{Chen2016} and the kernelized k-nearest-neighbor classifier (kknn)~\citep{Yu2002}. We decided to consider these learners because they represent three very distinct learning paradigms, and because they are generally regarded responsive to hyperparameter tuning. The hyperparameter spaces that are being tuned over are presented in Table~\ref{tab:hyperparams}. They also react differently to high dimensionality: kknn is vulnerable to the ``curse of dimensionality'', the SVM is a regularized modeling algorithm, and the xgboost algorithm does implicit feature selection because it is a tree-based learner.


\begin{table}
\caption{Hyperparameter spaces over which tuning was performed for the three learning algorithms \emph{xgboost}, \emph{support vector machine} and \emph{kernelized k-Nearest-Neighors} respectively. Ranges marked with~$^*$ were tuned on a logarithmic scale. For the large datasets madeline, madelon, arcene, AP\_Breast\_Colon and AP\_Colon\_Kidney, the xgboost \emph{nrounds} parameter was set fixed to 2000 and early stopping after 10 rounds was enabled. }
\label{tab:hyperparams}
\centering
\begin{tabular}{@{}llllll@{}}\toprule
\multicolumn{2}{c}{\textrm{SVM}} & ~~ & \multicolumn{2}{c}{\textrm{xgboost}} \\
\cmidrule{1-2} \cmidrule{4-5}
kernel & \textrm{rbfdot} & & \textrm{nrounds} & $\{1, 2, \ldots, 2000 \}$\\
\textrm{sigma} & $[2^{-10}, 2^{10}]~^*$ & &  \textrm{eta} & $[0.01, 0.2]$\\
\textrm{C} & $[2^{-10}, 2^{10}]~^*$ & & \textrm{gamma} & $[2^{-7}, 2^{6}]~^*$\\
& & & \textrm{max\textunderscore depth} & $\{3, \ldots, 20\}$\\
\multicolumn{2}{c}{\textrm{kknn}} & & \textrm{colsample\textunderscore bytree} & $[0.5, 1]$\\
\cmidrule{1-2}
\textrm{k} & $\{1, 2, \ldots, 50\}$ & & \textrm{colsample\textunderscore bylevel} &  $[0.5, 1]$\\
\textrm{distance} & $[1, 100]$ & & \textrm{lambda} & $[2^{-10}, 2^{10}]~^*$ \\
\textrm{kernel} & \{rectangular, & & \textrm{alpha} & $[2^{-10}, 2^{10}]~^*$   \\
& optimal, & & \textrm{subsample} & $[0.5, 1]$  \\
& triangular, & &  \\
& biweight\} & &  \\
\bottomrule
\end{tabular}
\end{table}

\subsection{Algorithms}\label{sec:algorithms}


We perform hyperparameter optimization and feature selection for the learners described in Section~\ref{sec:algos}. The datasets used are presented in Section~\ref{sec:tasks}. We use different configurations of our methods for comparisons described in the following. See also Figure~\ref{fig:indivexplanation} for a schematic representation of individual configuration vectors used in each method.

To answer research question~\ref{enum:gavsbo}, comparing the evolutionary and Bayesian optimization methods, we consider the following algorithms:
\begin{description}
    \item[GA-MO-FE] The NSGA2 with filter ensemble based initialization (as in~\ref{sec:gafi}) \emph{and} mutation (as in~\ref{sec:gafe}).
    \item[BO-MO-FE] Multi-objective Bayesian optimization with filter ensemble selection as described in~\ref{sec:bofe}.
\end{description}

To study research question~\ref{enum:fevsno}, i.e. the contribution of filter ensembles to the methods, we compare both methods to these algorithms without filter ensembles:
\begin{description}
    \item[GA-MO] The NSGA2 with filter ensemble based initialization (as in~\ref{sec:gafi}) but \emph{without} filter-based mutation, instead only using Hamming-weight preserving mutation as in~\ref{sec:ganofe}.
    \item[BO-MO] Multi-objective Bayesian optimization with individual filter selection, as described in~\ref{sec:bonofe}.
\end{description}

We illuminate research question~\ref{enum:movsso} by considering feature selection and hyperparameter optimization as a single-objective task, run with the following methods:
\begin{description}
    \item[BO-SO] Single-objective Bayesian optimization with individual filter selection (see~\ref{sec:bonofe})---this is a method similar to auto-sklearn's approach~\citep{Feurer2015} which we consider to be a state-of-the-art approach.
    \item[BO-SO-FE] Single-objective Bayesian optimization with filter ensemble selection (see~\ref{sec:bofe}), which is our strongest single objective baseline.
\end{description}

To address research question~\ref{enum:jvsnj}, we consider variants of our algorithms that perform feature selection and hyperparameter optimization in separate steps. We construct two algorithms to approximate the most straightforward simplifications to non-joint optimization. Note that the GA-approach has a focus on feature selection, while the BO-approach is more directed at hyperparameter tuning.
\begin{description}
    \item[GA-MO-FE-NJ] We initially optimize hyperparameters for the learning algorithms on each dataset by running standard single objective Bayesian optimization on the \emph{full} datasets without performing feature selection, through 500 model evaluations. The resulting hyperparameter values are then fixed, while our full NSGA2-variant (using both filter ensemble initialization and mutation) only performs multi-objective feature selection.
    \item[BO-MO-FE-NJ] To optimize hyperparameters and filter weights we use the BO-SO-FE method, and then evaluate the final model with the fixed hyperparameters. We iterate over equally spaced $\mathrm{ffrac}$ values and evaluate the model with all filters. This constructs a set of models with different trade-offs between sparsity and model performance, from which we construct a Pareto-set.
\end{description}

The methods have access to $M = 5$ filter methods\footnote{The methods were chosen by running many filters on a range of datasets and performing hierarchical clustering on the differences of their feature rankings; see the Supplement for more details.}: Information Gain, Random Forest Feature Importance~\citep{Ishwaran2010}, Joint Mutual Information (JMI)~\citep{Yang}, Minimal Conditional Mutual Information Maximization (CMIM)~\citep{Fleuret2004}, Area Under the Curve (AUC).

All NSGA2 variants use $\mu = 80$ as population and $\lambda = 15$ as offspring size which \citet{Khan2015} found to perform well in a feature selection setting. We choose an overall per-individual mutation probability of 0.3 and an overall per-pair-of-individuals crossover probability of 0.7.

Our Bayesian optimization methods use a random forest as surrogate to model the mixed discrete and continuous hyperparameter space similar to the state-of-the-art hyperparameter optimization toolbox SMAC~\citep{Huttera}. The infill criterion used is LCB~\citep{cox1992}. In each iteration, a batch of $15$ configurations is proposed in parallel as described in \citet{Horn}.

\subsection{Evaluation}

We measure the performance of resulting models by their mean misclassification error ($\mathrm{mmce} = \frac{1}{n}\sum_i \mathbb{I}_{ y^{(i)} \ne \hat y^{(i)}}$) on a validation set with ground truth values $y^{(i)}$ and model predictions $\hat{y}^{(i)}$. 


To get an estimate of the optimization performance that is unbiased by potential overtuning, we performed \emph{nested resampling}: During the whole optimization run, each optimization algorithm is only allowed to assess model performance $\mathrm{mmce}_\mathrm{optim}$ through (\emph{inner}) cross-validation on a optimization set $\mathcal{D}_{\mathrm{optim}}$. The final performance of the algorithm is reported as the performance of the solution candidates trained on $\mathcal{D}_{\mathrm{optim}}$ and evaluated on a test set $\mathcal{D}_{\mathrm{test}}$. This procedure is repeated $10$ times on \emph{outer} cross-validation folds $\left(\mathcal{D}^{(k)}_{\mathrm{optim}}, \mathcal{D}^{(k)}_{\mathrm{test}}\right), k = 1, ..., 10$.


As a proper multi-objective performance measure we consider the \emph{dominated hypervolume}~\citep{zitzler98} (\textrm{domHV}) with reference point $W = (1, 1)$, which corresponds to the worst possible values w.r.t.\ the two objectives. To prevent overtuning effects from skewing our results, we proceed as follows:

\begin{enumerate}
    \item We let each optimization algorithm report its Pareto set, taking into account only $\mathcal{D}_{\mathrm{optim}}$. This corresponds to returning non-dominated individuals w.r.t.\ $(\mathrm{mmce}_\mathrm{optim}, \mathrm{ffrac})$. GA-based methods take only candidates from their current generation, while BO-based methods get the Pareto set of all candidates seen so far.
    \item We calculate the \emph{generalization} dominated hypervolume $\textrm{domHV}_\mathrm{gen}$, i.e.\ the hypervolume that is dominated by this Pareto set of candidate configurations w.r.t.\ the $(\mathrm{mmce}_\mathrm{test}, \mathrm{ffrac})$ evaluated on $\mathcal{D}_{\mathrm{test}}$.\footnote{Note that $\textrm{domHV}_\mathrm{gen}$ is not the same as the dominated hypervolume of a population on the test set. Instead, only the individuals that are non-dominated according to $\mathcal{D}_{\mathrm{optim}}$ are used to calculate their dominated hypervolume on $\mathcal{D}_{\mathrm{test}}$.}
\end{enumerate}

For each of the experiments, we allow a maximum number of $2000$ model evaluations, with one evaluation corresponding to computing a single inner $10$-fold cross-validation on $\mathcal{D}_\mathrm{optim}$. This corresponds to approximately $130$ iterations for both the model-based and the evolutionary approach.

\section{Results}\label{sec:results}

\subsection{Evolutionary vs. Bayesian Optimization, Effect of Filter Ensembles}

\begin{figure}
\centering
\includegraphics[width=0.6\textwidth]{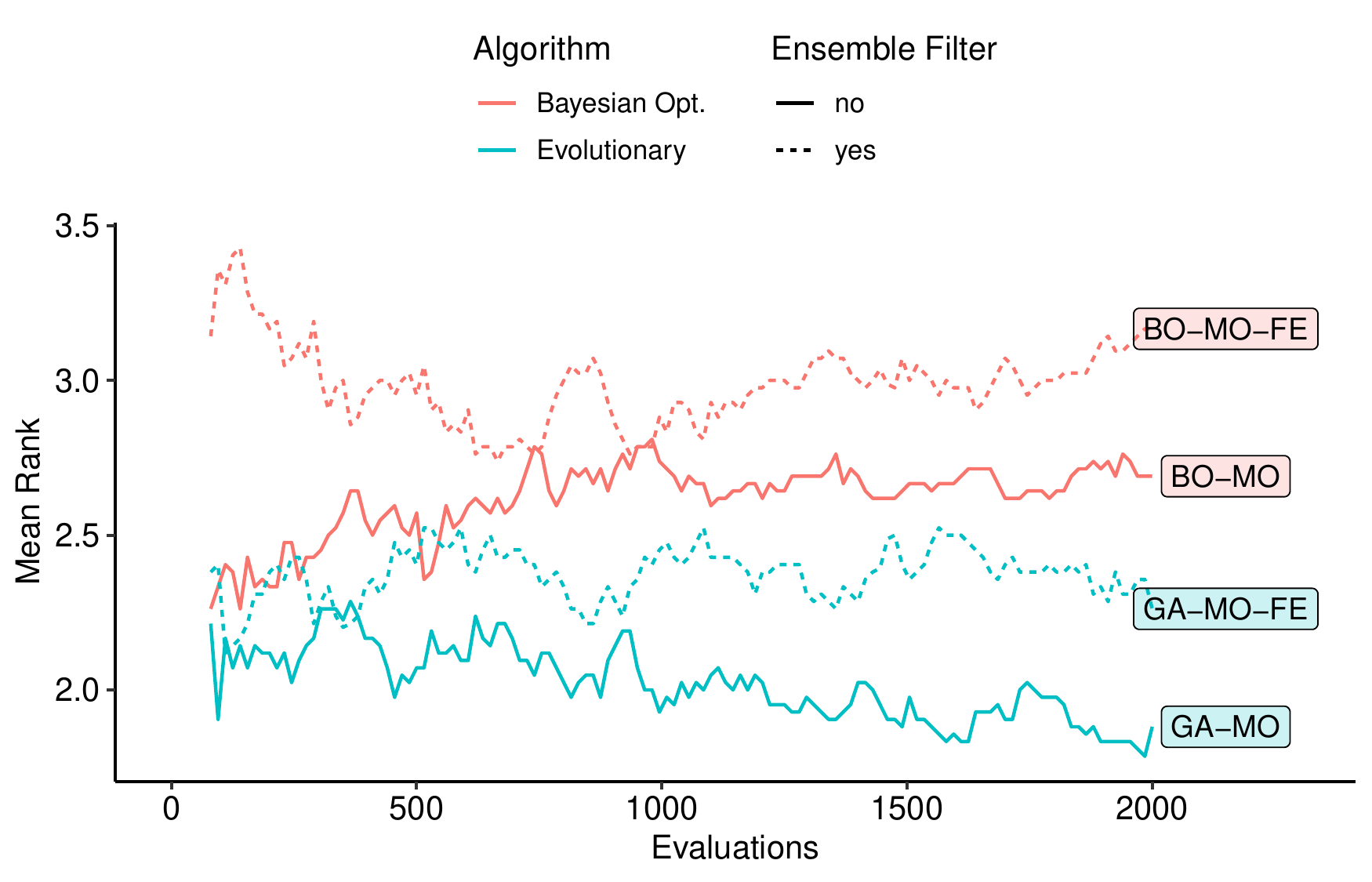} \\
\includegraphics[width=0.6\textwidth]{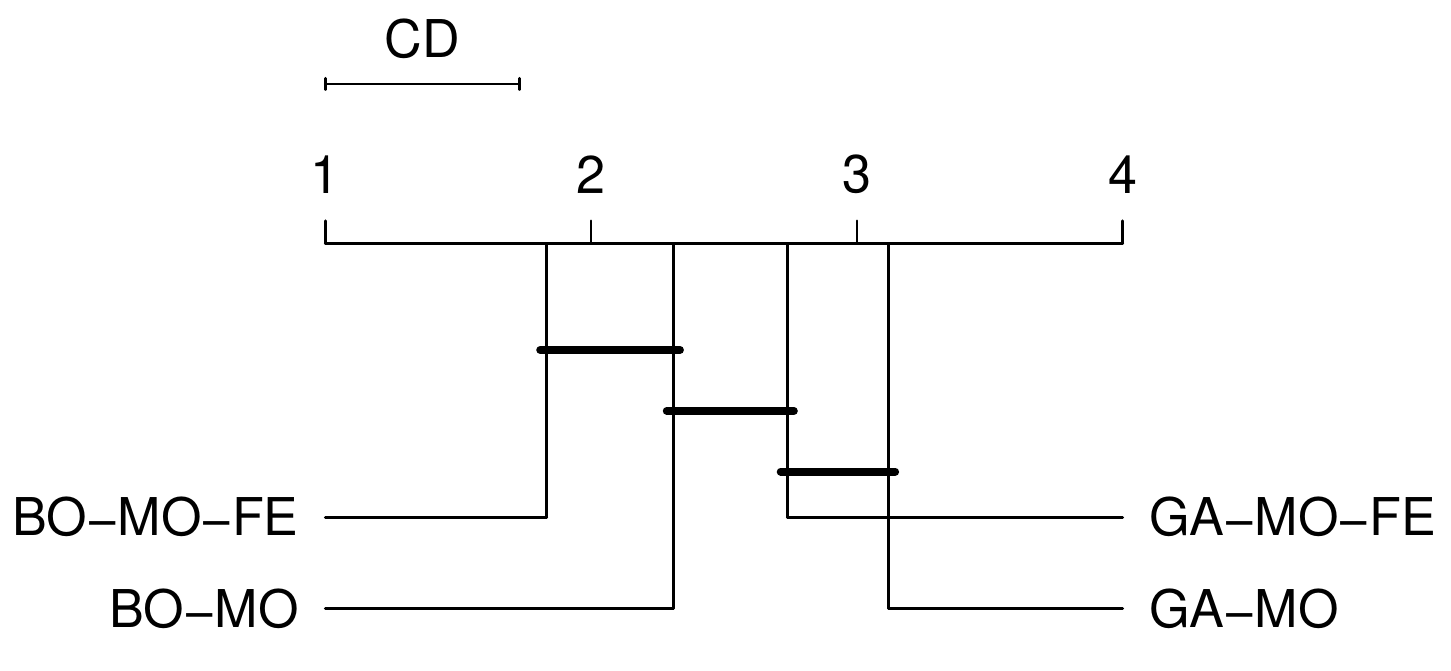}
\caption{Comparison of different multi-objective optimization methods. Top: the results of a global rank analysis based on $\textrm{domHV}_\mathrm{gen}$ to compare the proposed methods BO-MO(-FE) and GA-MO(-FE), i.e.\ both the BO and GA with and without filter ensemble. Ranks are computed per dataset and algorithm (ties are ranked by their average rank) and then averaged. Higher values are better. Bottom: Non-parametric critical difference test~\citep{demsar} performed after the full budget of 2000 evaluations at significance level $\alpha = 0.05$, based on $\textrm{domHV}_\mathrm{gen}$ rank as above. The BO-MO-FE method statistically significantly outperforms all other shown methods, although the absolute difference is small (see Figure~\ref{fig:runtime}).}
\label{fig:rq12}
\end{figure}

The global rank analysis as well as a critical difference test~\citep{demsar} presented in Figure~\ref{fig:rq12} show that our most advanced BO-based method, BO-MO-FE, significantly outperforms all GA-based methods. The optimization trace seems to indicate that including the feature ensemble also confers an advantage, although this is not statistically significant after 2000 evaluations.

\begin{figure}
\centering
\includegraphics[width=0.6\textwidth]{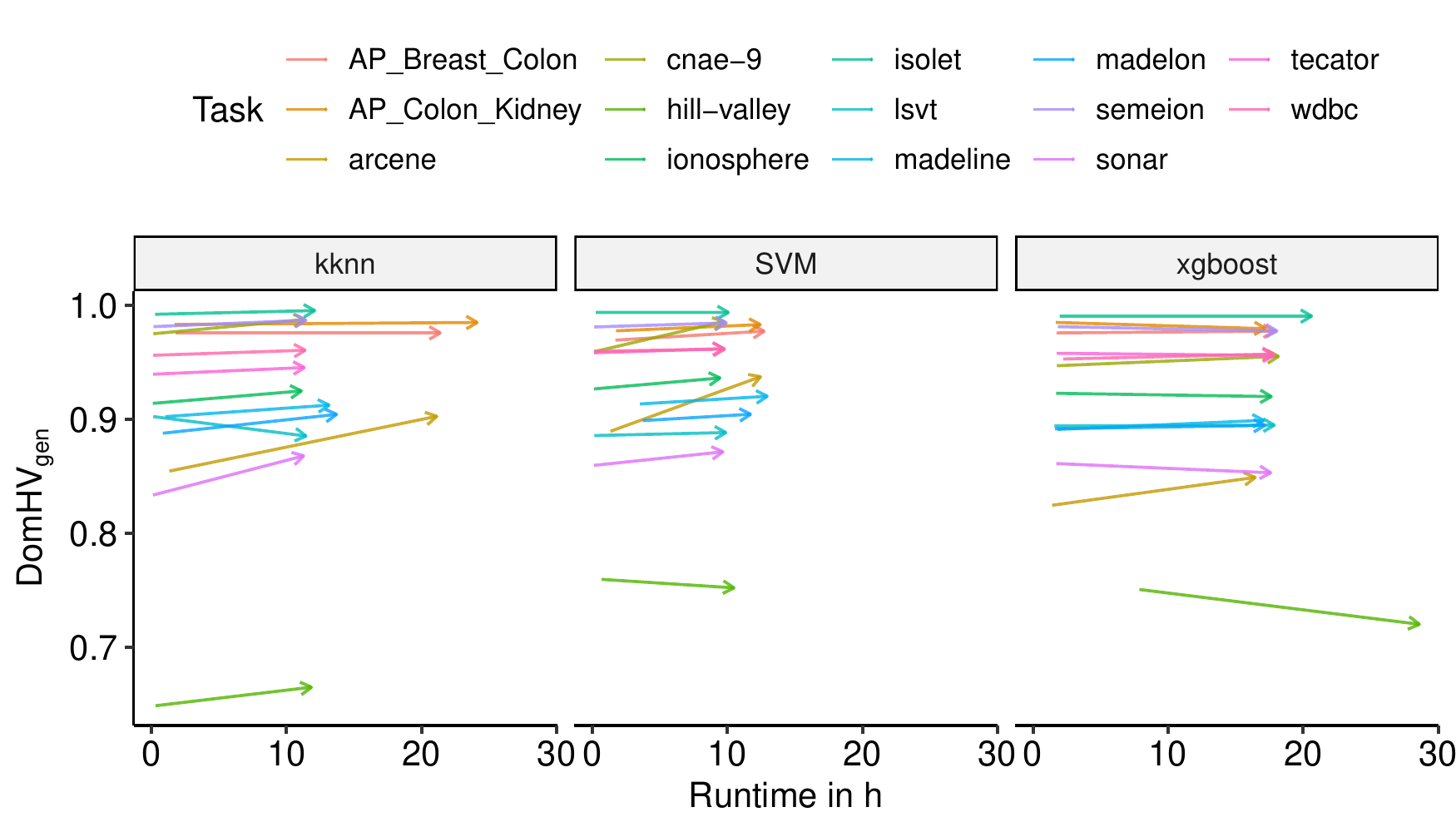} \\
\caption{Performance ($\textrm{domHV}_\mathrm{gen}$) and runtime of GA-MO-FE (tail end of arrow) and BO-MO-FE (head of arrow) on each dataset (Table~\ref{tab:datasets}) and learning algorithm (Table~\ref{tab:hyperparams}), averaged over 10 outer cross-validation runs. BO-MO-FE has moderately improved performance over GA-MO-FE, but at a cost in runtime.}
\label{fig:runtime}
\end{figure}

Figure~\ref{fig:runtime} shows an additional aspect to take into consideration: overall runtime. BO-MO-FE usually has a slight performance advantage over GA-MO-FE in absolute terms, but comes with computational overhead that may have to be considered if individual model performance evaluations are relatively fast. The overhead becomes less relevant when optimizing large datasets or slow models.

\subsection{Multi-Objective vs. Single-Objective}\label{sec:moso}
\begin{figure}
\centering
\includegraphics[width=0.6\textwidth]{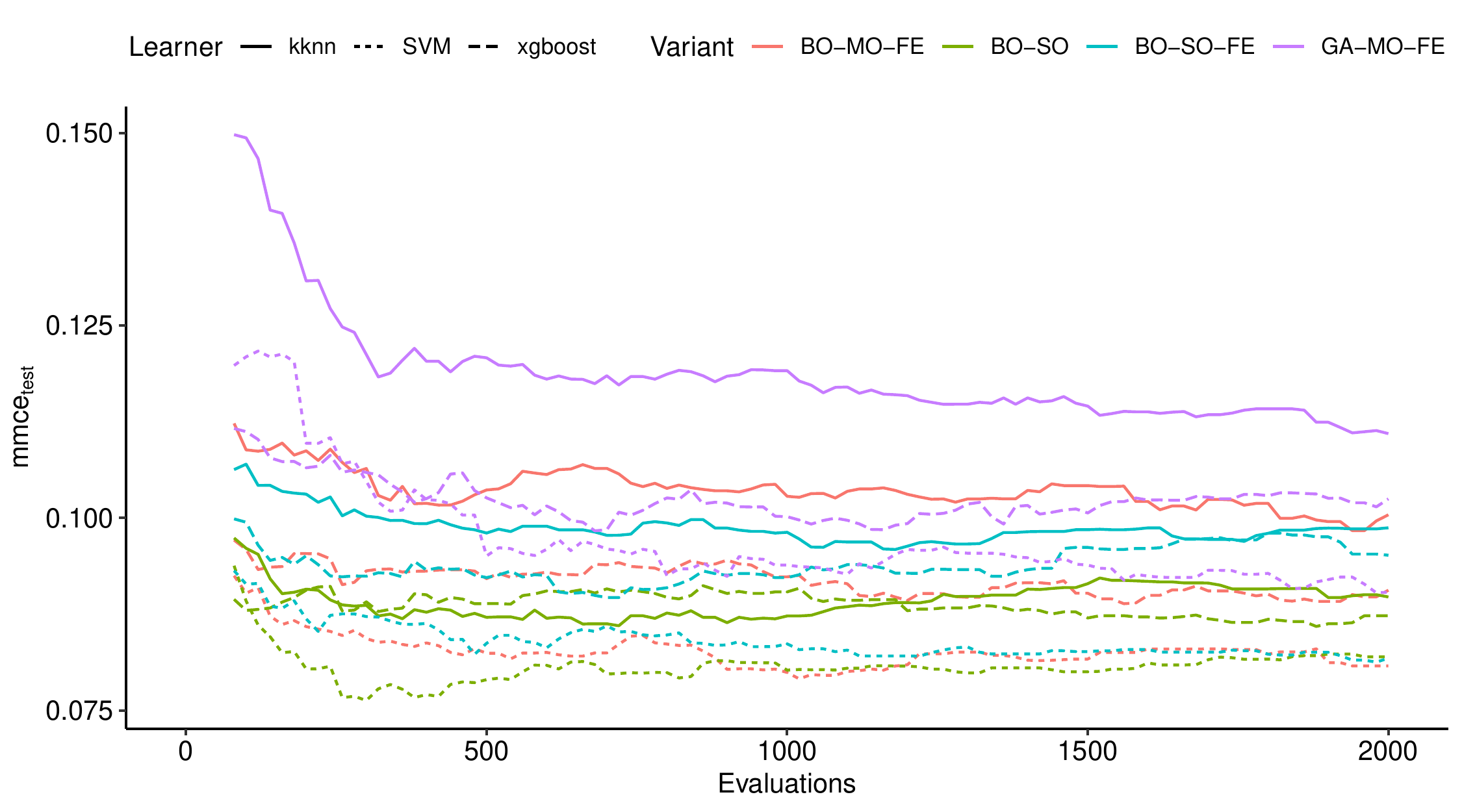} \\
\caption{Comparison of multi-objective and single-objective methods. Shown is the $\mathrm{mmce}_\mathrm{test}$ of the individual with best $\mathrm{mmce}_\mathrm{optim}$ found until a particular evaluation, averaged over 10 outer cross-validation folds and over all datasets. Lower values are better. Note that $\mathrm{mmce}_\mathrm{test}$ is not monotonically improving because individuals with better $\mathrm{mmce}_\mathrm{optim}$ may be found that perform worse on the test set than previous individuals.}
\label{fig:soperf}
\end{figure}

The goal of our multi-objective methods is not to optimize predictive performance by itself, and instead to explore the possible trade-offs between performance and sparseness. It is still interesting to look at the best performing model configurations being found by the multi-objective methods, and to compare them to the best models found by single-objective methods, as done in Figure~\ref{fig:soperf}. Here, again, the BO-based methods outperform GA-based methods. Except for the \emph{kknn} learning algorithm, the MO-method is on par with the SO methods.

\begin{figure}
\centering
\includegraphics[width=0.6\textwidth]{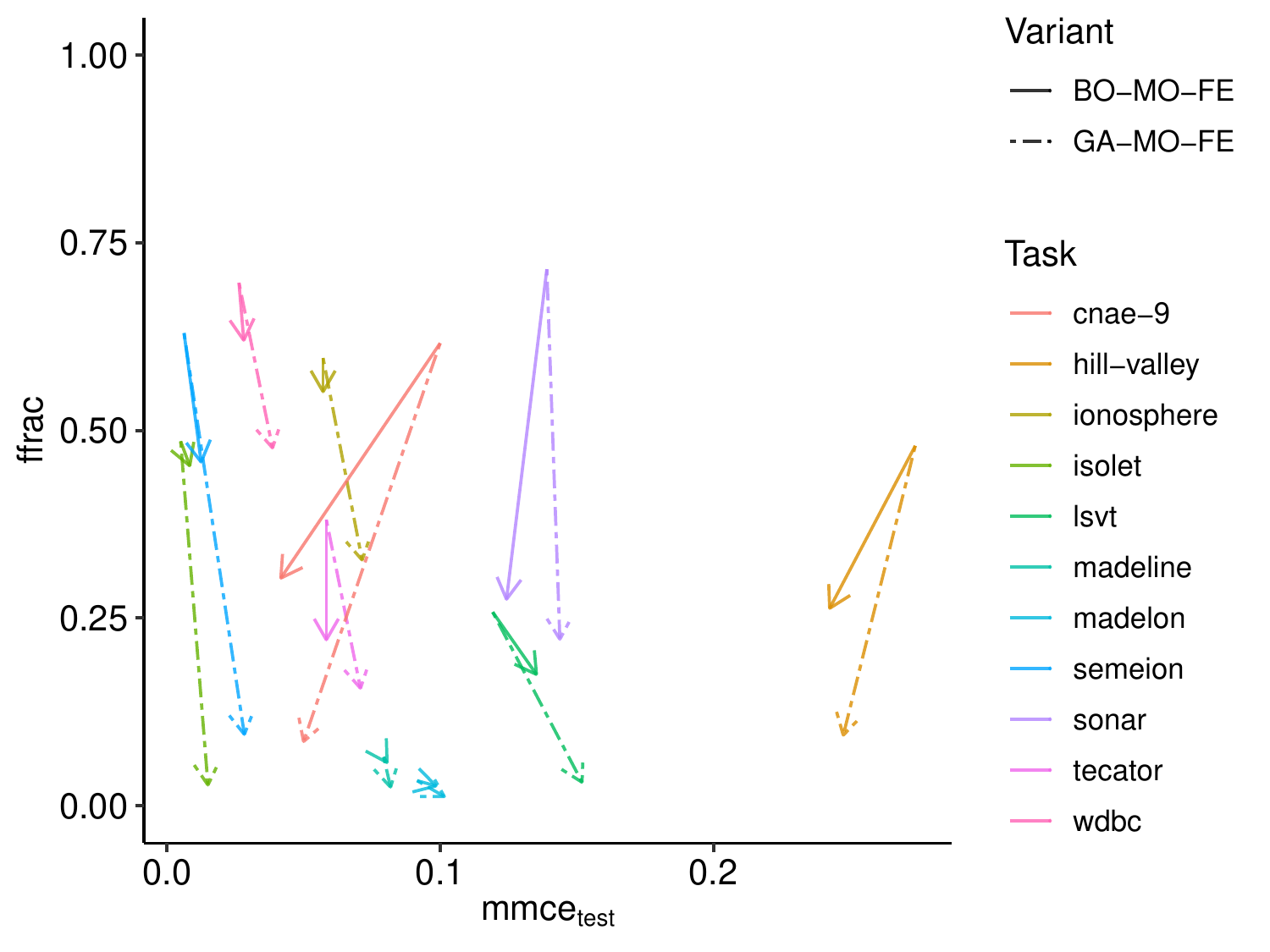} \\
\caption{Comparison of multi-objective and single-objective methods applied to the SVM learning algorithm: Performance $\mathrm{mmce}_\mathrm{test}$ and the fraction of included features $\mathrm{ffrac}$ found after 2000 evaluations by baseline BO-SO (tail end of arrows), BO-MO-FE (head of solid arrows) and GA-MO-FE (head of dashed arrows). Each dataset (Table~\ref{tab:datasets}) is shown, values are averaged over 10 outer CV runs. Choice of individuals for MO methods described in Section~\ref{sec:moso}.}
\label{fig:soarrows}
\end{figure}

The BO-SO method employs a feature filtering step, and the models it chooses will often be sparse to some degree, for regularization. Figure~\ref{fig:soarrows} compares the BO-SO method to the multi-objective methods in terms of both sparseness and predictive performance: For each learning algorithm, dataset, and cross-validation fold of BO-SO, the best performing model from BO-MO-FE and GA-MO-FE were chosen that are \emph{at least} as sparse. The plot shows that both MO methods often found much sparser models than the SO baseline while giving up very little in predictive performance.

\subsection{Simultaneous Hyperparameter Tuning and Feature Selection}
Table~\ref{tab:nonjoint-results} shows the $\textrm{domHV}_\mathrm{gen}$ of our joint optimization methods, compared to their non-joint correspondents. In most cases the joint optimization confers a considerable performance advantage, especially compared to the BO-SO-FE-NJ method. We assume that the advantage of joint over non-joint methods depends on how much the hyperparameter performance of a model interacts with sparseness.

\begin{table}
    \small
    \centering
    \caption{$\textrm{domHV}_\mathrm{gen}$ after $2000$ evaluations of simultaneous hyperparameter tuning / feature selection methods compared to corresponding non-simultaneous methods. Best results for GA and BO in bold. Values averaged over datasets, see supplement for results by dataset.
    }
    \label{tab:nonjoint-results}
    \begin{tabular}{lrrrr}
      \toprule
Learner & BO-MO-FE & BO-MO-FE-NJ & GA-MO-FE & GA-MO-FE-NJ \\ 
  \hline
kknn & \textbf{0.9217} & 0.8670 & 0.9108 & \textbf{0.9134} \\ 
SVM & \textbf{0.9331} & 0.8719 & \textbf{0.9241} & 0.8892 \\ 
xgboost & \textbf{0.9164} & 0.8940 & \textbf{0.9165} & 0.9121 \\ 
       \bottomrule
    \end{tabular}
\end{table}

\section{Conclusion}\label{sec:conclusion}

Our results show that both evolutionary approaches and model-based approaches can efficiently perform model-agnostic multi-objective optimization to simultaneously tune hyperparameters and select features. We have also shown that performing these tasks simultaneously has an advantage over running them separately.

Using model-based optimization to tune over the type of filter measures as well as fraction of included features is used in some AutoML-frameworks~\citep{Feurer2015}. Our experiments suggest that the adaption of this to multi-objective optimization works well, although using a parameterized filter ensemble outperforms it.

Our other proposed method based on an NSGA-II enhanced with specialized initialization and mutation sampling seems to perform almost as well, although it does not reach the level of the best Bayesian optimization based approach. The advantage of the NSGA-II, however, is that it does not introduce as much computational overhead, which can make up a considerable part of overall runtime if model performance evaluations themselves are cheap.

Our final recommendation is therefore to use the Bayesian optimization approach in combination with parameterized Filter ensembles if model evaluations are expensive, if computational resources are cheap, and if it is important to get configurations that perform very close to optimal, given their sparseness. The NSGA-II approach is suitable if model evaluations are cheap and if marginal degradation of performance are acceptable.

\section*{Acknowledgments}

This work has been partially funded by the German Federal Ministry of Education and Research (BMBF) under Grant No. 01IS18036A. The authors of this work take full responsibilities for its content.

This work was supported by the Bavarian Ministry for Economic Affairs, Infrastructure, Transport and Technology through the Center for Analytics-Data-Applications (ADA-Center) within the framework of “BAYERN DIGITAL II”.

\newpage

\bibliography{refs} 

\newpage
\onecolumn

\section*{Supplementary Material}

\begin{appendix}

\begin{figure*}[b!]
    \centering
    \includegraphics[width=0.9\textwidth]{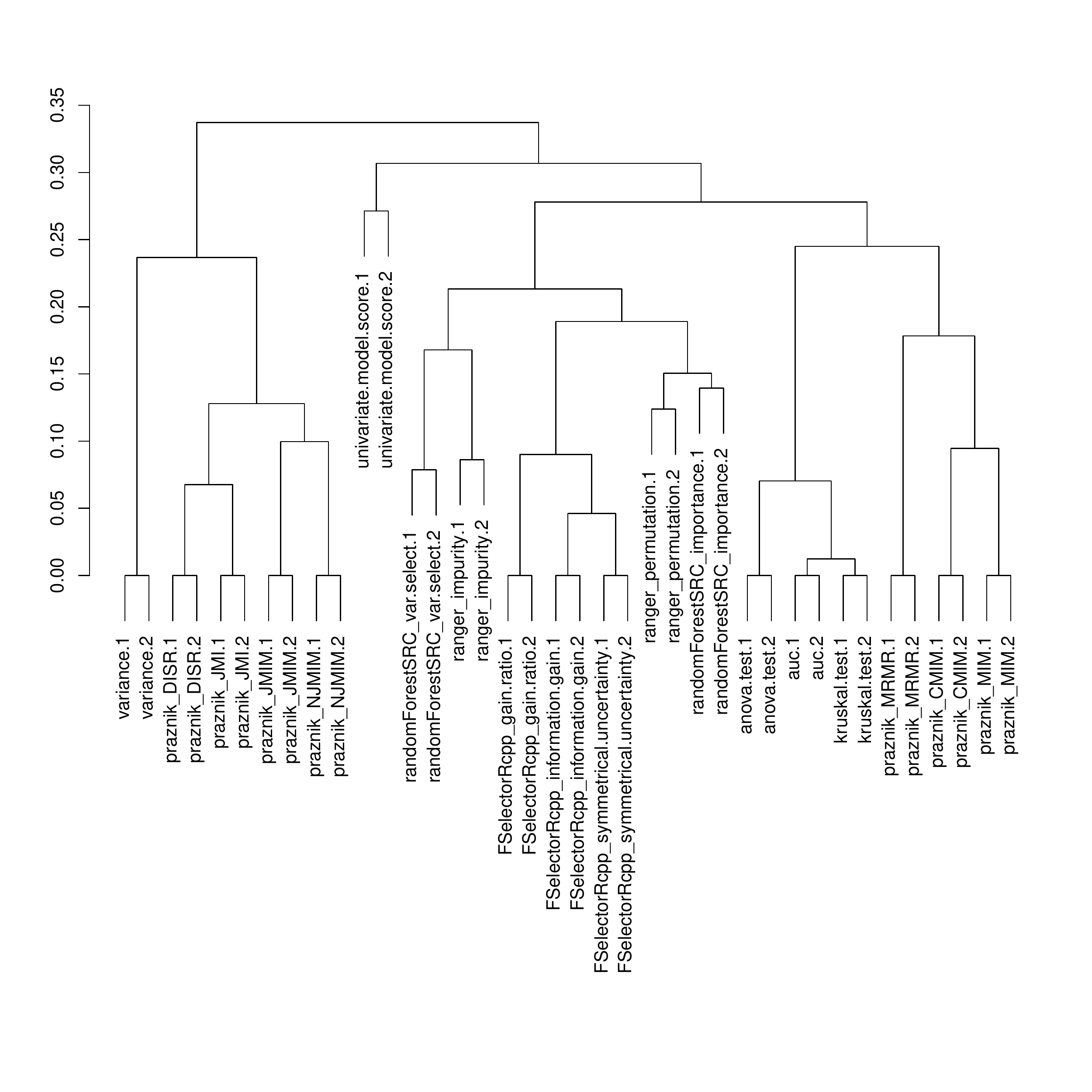}
    \caption{Cluster-dendrogram of filter values. We used a range of filter scoring methods included in the \texttt{mlr} R-package (\url{https://cran.r-project.org/web/packages/mlr/index.html})
    and evaluated each of them twice on a range of datasets. Distances between filters was calculated as the average (over datasets) L1-distance between filter-values that were rank-transformed and scaled to $[0, 1]$. The vertical axis gives the distance between clusters. For deterministic filter methods, both evaluation instances have distance $0$. The two instances of \textrm{univariate.model.score} differ more from each other than many other groups, so this method was excluded.}
    \label{fig:cluster}
\end{figure*}

\newpage

\subsection*{NSGA-II Ablation Study}

\begin{figure}[b]
    \centering
    
    \includegraphics[width=0.6\textwidth]{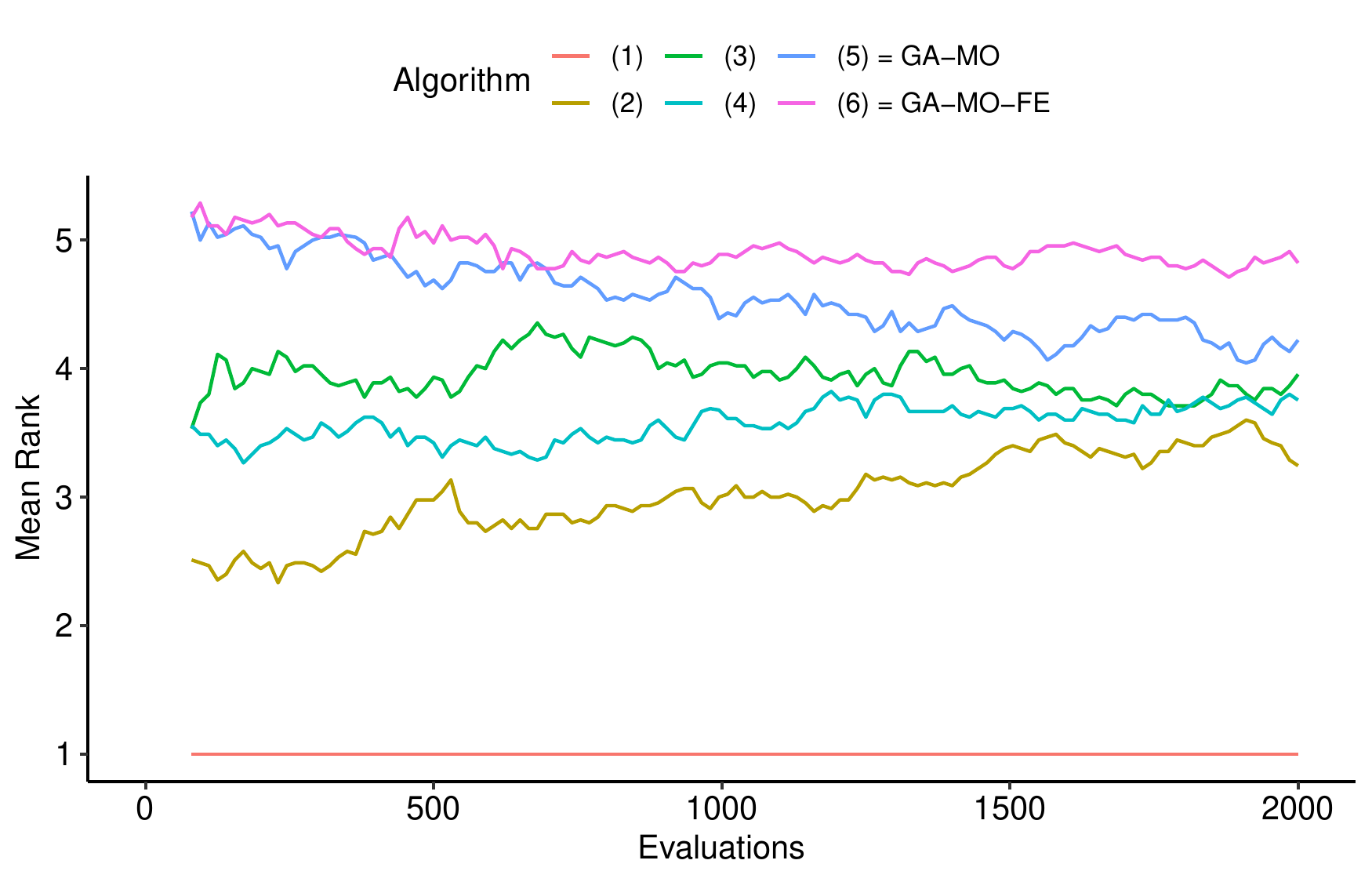}
    \small
    \caption{Global rank analysis to study the effect of the different algorithmic components (Table~\ref{tab:ablation}) of the GA based on $\textrm{domHV}_\mathrm{gen}$. Ranks are computed per dataset and algorithm (ties are ranked by their average rank) and then averaged. Higher values are better.}    
    \label{fig:ablation}
\end{figure}

We show that our methods introduced in Section~\ref{sec:methods} have beneficial effect on optimization performance. For this we perform an ablation study of running a basic NSGA-II, as well as different versions that have incrementally more specialized operations included. Figure~\ref{fig:ablation} shows the performance of different methods that successively include uniformly distributed and geometric initialization (Section~\ref{sec:geominit}), Hamming-weight preserving mutation (Section~\ref{sec:ganofe}), and Filter-ensemble based initialization and mutation (Sections \ref{sec:gafi} and \ref{sec:gafe}). The basic algorithm, closely corresponding to the method shown in \citet{Bouraoui2018}, has performance far inferior to all GA methods because its initial population does not cover the $\mathrm{ffrac}$ dimension well (Figure~\ref{fig:geominit}).

\begin{table}[b]
\centering
\caption{Description of different GA runs shown in Figure~\ref{fig:ablation}. ``Naive'' Bernoulli initialization consists of drawing each feature selection bit independently from a $\frac{1}{2}$-Bernoulli distribution.}
\label{tab:ablation}
\begin{tabular}{lcccc}
    \toprule
    Variant & Initialization & Hamming-weight & Filter-ensemble &  Filter-ensemble\\
           & (Section~\ref{sec:geominit}) & preserving mutation & initialization (Section~\ref{sec:gafi}) & mutation  \\
           & & (Section~\ref{sec:ganofe}) & &  (Section~\ref{sec:gafe}) \\
     \hline
    (1) & Bernoulli (naive) & No & No & No  \\
    (2) & uniform & No & No & No  \\
    (3)& geom & No & No & No \\
    (4)& geom & Yes & No & No \\
    (5) (GA-MO) & geom & Yes & Yes & No \\
    (6) (GA-MO-FE) & geom & Yes & Yes & Yes \\
    \bottomrule 
    \end{tabular}
\end{table}

\begin{figure}
\centering
\includegraphics[width=0.6\textwidth]{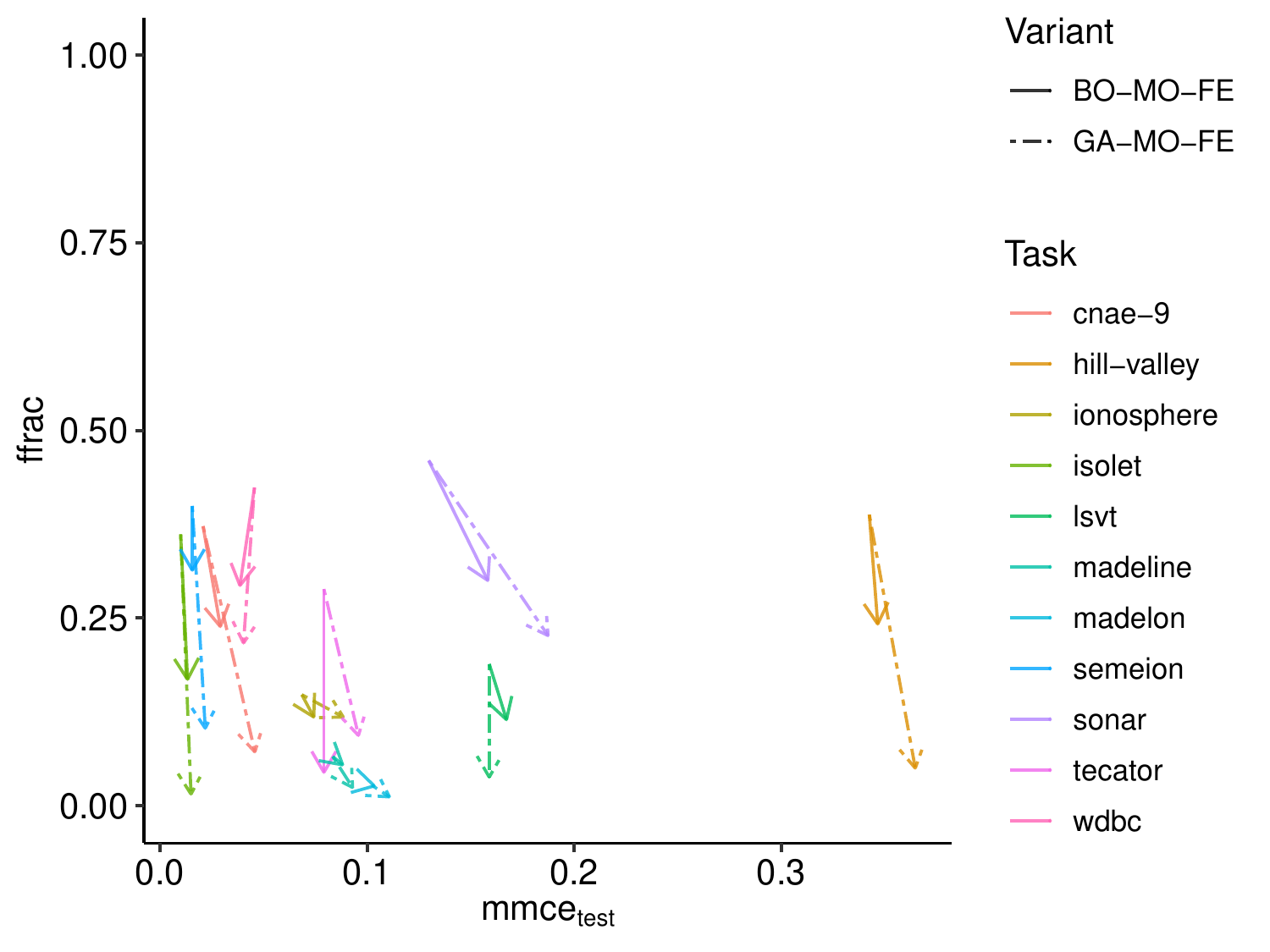} \\
\caption{Comparison of multi-objective and single-objective methods applied to the kknn learning algorithm. Shown is both the performance ($\mathrm{mmce}_\mathrm{test}$) and the fraction of included features ($\mathrm{ffrac}$) of individuals found after 2000 evaluations by the single-objective baseline BO-SO (tail end of arrows), BO-MO-FE (head of solid arrows) and GA-MO-FE (head of dashed arrows). Each dataset (Table~\ref{tab:datasets}) is shown, values are averaged over 10 outer cross-validation runs. Individuals of MO methods were chosen as described in Section~\ref{sec:moso}.}
\label{fig:soarrows-kknn}
\end{figure}

\begin{figure}
\centering
\includegraphics[width=0.6\textwidth]{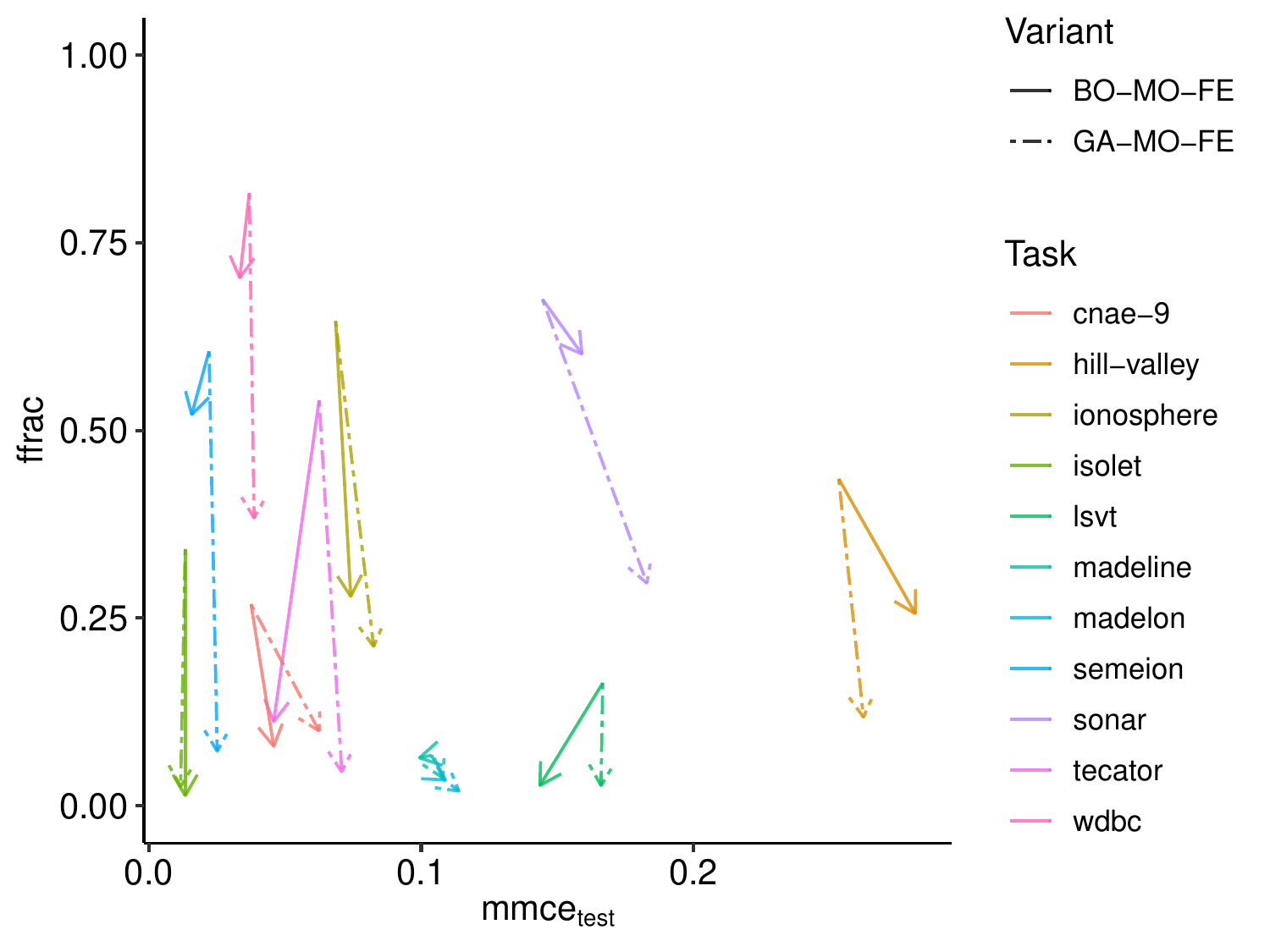} \\
\caption{Comparison of multi-objective and single-objective methods applied to the xgboost learning algorithm. Shown is both the performance $\mathrm{mmce}_\mathrm{test}$ and the fraction of included features $\mathrm{ffrac}$ of individuals found after 2000 evaluations by the single-objective baseline BO-SO (tail end of arrows), BO-MO-FE (head of solid arrows) and GA-MO-FE (head of dashed arrows). Each dataset (Table~\ref{tab:datasets}) is shown, values are averaged over 10 outer cross-validation runs. Individuals of MO methods were chosen as described in Section~\ref{sec:moso}.}
\label{fig:soarrows-xgboost}
\end{figure}

\begin{table}
\centering
    \caption{Methods that perform hyperparameter tuning and feature selection simultaneously (BO-MO-FE, GA-MO-FE) are compared to similar methods that do not perform the tasks simultaneously. The table shows the $\textrm{domHV}_\mathrm{gen}$ after $2000$ evaluations.}
    \label{tab:nonjoint-results-detailed}
    \begin{tabular}{llrrrrr}
      \toprule
& Learner & Problem & BO-MO-FE & BO-MO-FE-NJ & GA-MO-FE & GA-MO-FE-NJ \\ 
  \hline
1 & kknn & AP\_Breast\_Colon & 0.9761 & {0.9849} & 0.9761 & 0.9761 \\ 
  2 & kknn & AP\_Colon\_Kidney &{0.9853} & 0.9595 & 0.9836 & {0.9872} \\ 
  3 & kknn & arcene & 0.9030 & 0.8382 & 0.8549 & 0.8898 \\ 
  4 & kknn & cnae-9 & 0.9879 & 0.9692 & 0.9754 & 0.9662 \\ 
  5 & kknn & hill-valley & 0.6653 & 0.6548 & 0.6490 & 0.6722 \\ 
  6 & kknn & ionosphere & 0.9252 & 0.8581 & 0.9143 & 0.8989 \\ 
  7 & kknn & isolet & 0.9957 & 0.9908 & 0.9925 & 0.9959 \\ 
  8 & kknn & lsvt & 0.8856 & 0.8675 & 0.9027 & 0.8770 \\ 
  9 & kknn & madeline & 0.9128 & 0.7505 & 0.9027 & 0.8894 \\ 
  10 & kknn & madelon & 0.9047 & 0.6201 & 0.8881 & 0.8686 \\ 
  11 & kknn & semeion & 0.9872 & 0.9606 & 0.9817 & 0.9805 \\ 
  12 & kknn & sonar & 0.8683 & 0.8374 & 0.8338 & 0.8881 \\ 
  13 & kknn & tecator & 0.9458 & 0.9162 & 0.9399 & 0.9368 \\ 
  14 & kknn & wdbc & 0.9610 & 0.9297 & 0.9565 & 0.9614 \\ 
  15 & SVM & AP\_Breast\_Colon & 0.9777 & 0.9401 & 0.9698 & 0.9777 \\ 
  16 & SVM & AP\_Colon\_Kidney & 0.9835 & 0.9595 & 0.9780 & 0.9107 \\ 
  17 & SVM & arcene & 0.9377 & 0.8732 & 0.8898 & 0.7199 \\ 
  18 & SVM & cnae-9 & 0.9862 & 0.9660 & 0.9597 & 0.9590 \\ 
  19 & SVM & hill-valley & 0.7525 & 0.7093 & 0.7599 & 0.7653 \\ 
  20 & SVM & ionosphere & 0.9366 & 0.9082 & 0.9270 & 0.9318 \\ 
  21 & SVM & isolet & 0.9941 & 0.9941 & 0.9942 & 0.9942 \\ 
  22 & SVM & lsvt & 0.8887 & 0.8863 & 0.8861 & 0.8821 \\ 
  23 & SVM & madeline & 0.9205 & 0.6468 & 0.9138 & 0.7789 \\ 
  24 & SVM & madelon & 0.9048 & 0.6219 & 0.8992 & 0.7389 \\ 
  25 & SVM & semeion & 0.9849 & 0.9756 & 0.9814 & 0.9817 \\ 
  26 & SVM & sonar & 0.8718 & 0.8442 & 0.8599 & 0.8847 \\ 
  27 & SVM & tecator & 0.9624 & 0.9469 & 0.9587 & 0.9633 \\ 
  28 & SVM & wdbc & 0.9619 & 0.9339 & 0.9600 & 0.9599 \\ 
  29 & xgboost & AP\_Breast\_Colon & 0.9777 & 0.9738 & 0.9761 & 0.9777 \\ 
  30 & xgboost & AP\_Colon\_Kidney & 0.9799 & 0.9800 & 0.9853 & 0.9781 \\ 
  31 & xgboost & arcene & 0.8494 & 0.8405 & 0.8249 & 0.8349 \\ 
  32 & xgboost & cnae-9 & 0.9555 & 0.9434 & 0.9473 & 0.9546 \\ 
  33 & xgboost & hill-valley & 0.7205 & 0.6932 & 0.7510 & 0.7390 \\ 
  34 & xgboost & ionosphere & 0.9203 & 0.8928 & 0.9232 & 0.9239 \\ 
  35 & xgboost & isolet & 0.9909 & 0.9877 & 0.9908 & 0.9892 \\ 
  36 & xgboost & lsvt & 0.8952 & 0.8376 & 0.8945 & 0.8559 \\ 
  37 & xgboost & madeline & 0.8994 & 0.8502 & 0.8915 & 0.8906 \\ 
  38 & xgboost & madelon & 0.8950 & 0.8405 & 0.8921 & 0.8847 \\ 
  39 & xgboost & semeion & 0.9778 & 0.9717 & 0.9817 & 0.9845 \\ 
  40 & xgboost & sonar & 0.8534 & 0.8245 & 0.8615 & 0.8378 \\ 
  41 & xgboost & tecator & 0.9564 & 0.9497 & 0.9583 & 0.9654 \\ 
  42 & xgboost & wdbc & 0.9576 & 0.9310 & 0.9533 & 0.9529 \\ 
       \bottomrule
    \end{tabular}
\end{table}

\end{appendix}

\end{document}